\documentclass{article}

\PassOptionsToPackage{numbers, compress}{natbib}
\usepackage[dblblindworkshop, final]{neurips_2026}

\usepackage[utf8]{inputenc}
\usepackage[T1]{fontenc}
\usepackage{xurl}       
\usepackage{hyperref}
\hypersetup{
  pdftitle={What Should World Models Forget? Stratified Retention for Continual Adaptation},
  pdfauthor={Nishit Anand, Ramani Duraiswami, Dinesh Manocha}
}
\usepackage{booktabs}
\usepackage{array}
\usepackage{tabularx}
\usepackage{amsfonts}
\usepackage{amssymb}
\usepackage{amsmath}
\usepackage{nicefrac}
\usepackage{microtype}
\usepackage{xcolor}
\usepackage{graphicx}

\title{What Should World Models Forget?\\ Stratified Retention for Continual Adaptation}

\workshoptitle{Continual World Models}

\author{%
  Nishit Anand \quad Ramani Duraiswami$^{\dagger}$ \quad Dinesh Manocha$^{\dagger}$ \\
  University of Maryland, College Park, USA \\
  \texttt{nishit@umd.edu}
}

\begin{document}

\maketitle
{\renewcommand{\thefootnote}{\fnsymbol{footnote}}\footnotetext[2]{Equal advising.}}

\begin{abstract}
Continual learning treats any drop in performance on previously seen data as forgetting, a convention inherited from settings where the prediction target is stationary. World models violate this assumption: their target is the environment, which changes, so knowledge that was accurate when learned can later become false, and revising it is correct behavior rather than a defect. Non-stationary ground truth is well studied in the concept drift literature and in the temporal factuality of language models, but it has not been formulated for world models, which are distinctive in that they also encode structural knowledge, such as physical laws and object permanence, that must never be degraded. We argue that continual world models need retention stratified by \emph{invariance timescale}, the time over which knowledge stays true: invariants should be refined but never degraded, while instance-level facts should be revised once the environment changes. We show that standard forgetting metrics, used alone, count correct revision as forgetting and so rank a frozen model highest. We propose \emph{differential retention}, which reports two quantities separately: invariant retention, checked after every adaptation round, and revision latency, the number of observations a model needs to pick up a change.
\end{abstract}

\section{Introduction}

A world model predicts how an environment evolves: given the current state $s_t$ and an action $a_t$, it predicts a distribution over the next state, $p(s_{t+1}\mid s_t,a_t)$. Agents use world models to plan, to train policies in imagination, and to generate experience that would be costly or unsafe to collect in the real world. Recent systems have scaled this idea considerably. Dreamer~4 learns to obtain diamonds in Minecraft from offline data alone \citep{hafner2025training}, and V-JEPA~2 pretrains on over one million hours of video before planning robot manipulation with an action-conditioned predictor in its latent space \citep{assran2025vjepa2}. In both cases, however, training ends before deployment, and the model is not updated from what it observes afterwards.

The environments these models operate in do not stay fixed. Continual learning provides tools for updating a model over time, but it rests on an assumption that world models do not satisfy: that the prediction target is stationary, so any decline in performance on previously learned data is a defect. This convention comes from supervised classification, where an image of a cat remains an image of a cat, and a model that becomes less accurate on it has truly degraded. A world model is in a different position. Suppose a household robot has learned that a chair stands by the window, and someone moves the chair to the table. The robot's old prediction is now false, and a model that keeps making it is stale rather than faithful, so its accuracy on the old data should drop.

This example separates three phenomena that current metrics cannot tell apart. \textbf{Catastrophic forgetting} is the loss of knowledge that is still true, such as forgetting where the unmoved table stands; it is a defect. \textbf{Deliberate unlearning} is the removal of knowledge that is still true but unwanted, for instance for privacy, and is well studied \citep{bourtoule2021machine}. \textbf{Obsolescence} is the retention of knowledge that has become false, such as still predicting the chair by the window; here, revision is required.

\begin{figure}[t]
\centering
\begin{minipage}[b]{0.50\linewidth}\raggedright
  {\footnotesize\bfseries (a) Protection should increase with $\tau$}\\[3pt]
  \includegraphics[width=\linewidth]{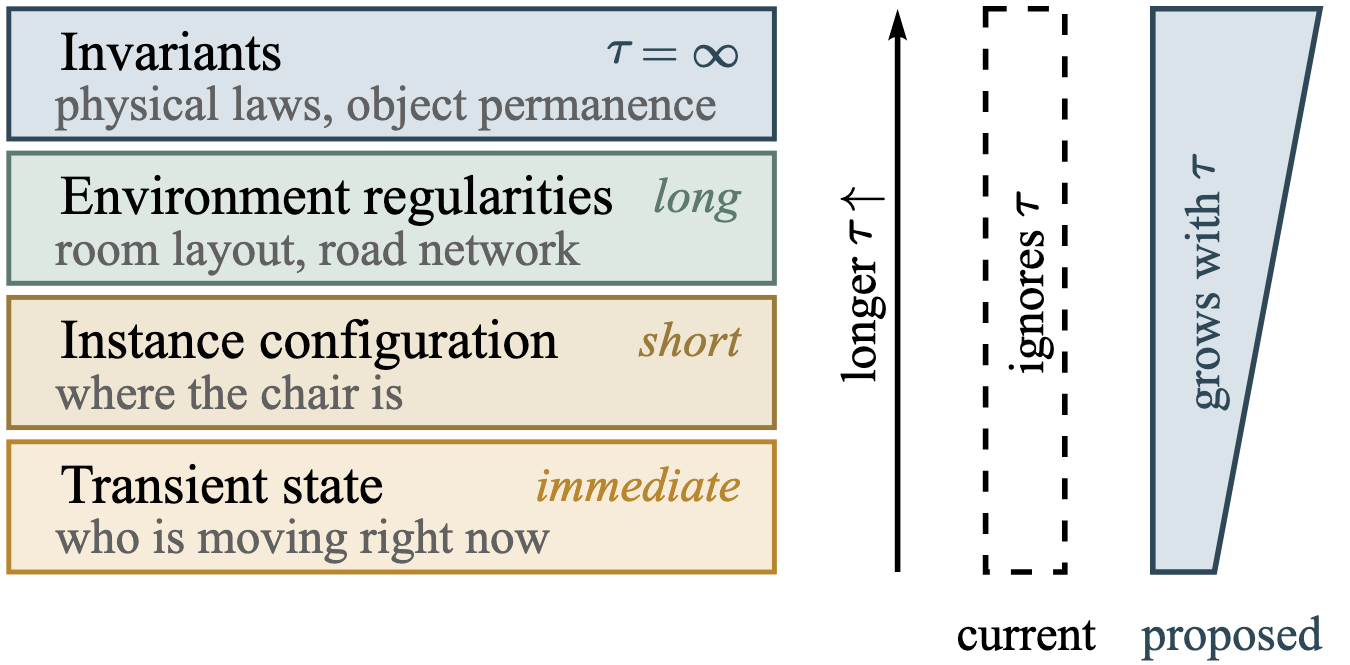}
\end{minipage}\hfill
\begin{minipage}[b]{0.365\linewidth}\raggedright
  {\footnotesize\bfseries (b) What the two measurements separate}\\[3pt]
  \includegraphics[width=\linewidth]{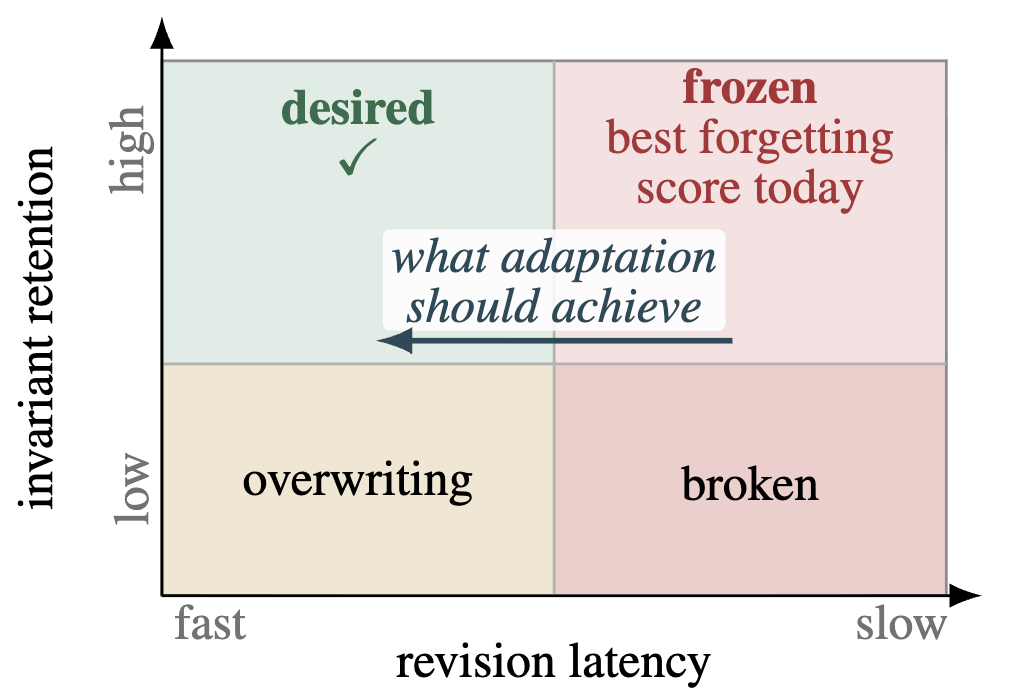}
\end{minipage}
\caption{(a) World-model knowledge in four strata of invariance timescale $\tau$ (Table~\ref{tab:strata}). Existing methods choose what to protect without regard to $\tau$ (dashed bar); we argue that protection, shown as bar width, should grow with $\tau$ (solid bar). (b) Invariant retention and revision latency together place a model in one of four quadrants; forgetting metrics alone reward high retention and slow revision, and so favor the frozen model.}
\label{fig:main}
\end{figure}

Changing ground truth is not new in machine learning. The concept drift literature assumes that the correct output for an input can change over time \citep{cabreramartin2026evolving}, and in language models, \citet{jiang2026benchmarks} measure how often an up-to-date answer is marked wrong by an outdated benchmark label. The problem has not, however, been posed for world models, and does not carry over unchanged. In both settings above, knowledge consists of discrete facts that can be checked and relabeled one at a time, including facts that never change \citep{jang2022towards}. A world model is different. Alongside facts that change, such as where the chair is, it encodes physical, geometric and causal structure, for example that unsupported objects fall, that solid objects do not pass through one another, and that hidden objects continue to exist. This structure should survive every update, it underlies every prediction the model makes, and no single label can check it. World models are therefore the setting in which obsolescence coexists with hard invariants, and this combination changes both what must be protected and what must be measured (Figure~\ref{fig:main}).

\textbf{Contributions.} We argue that obsolescence is a failure mode of continual world models that is distinct from catastrophic forgetting, and that current benchmarks exclude by construction (Sec.~\ref{sec:assumption}). We propose organizing world-model knowledge by its \emph{invariance timescale} $\tau$ and protecting knowledge more strongly the longer it stays true (Sec.~\ref{sec:strata}). We show that standard forgetting metrics cannot distinguish correct revision from damage (Sec.~\ref{sec:metrics}). Finally, we propose \emph{differential retention}, an evaluation protocol that measures invariant retention and revision latency separately (Sec.~\ref{sec:protocol}). The appendix covers related work, the strata, the protocol and the probes in detail.

\section{World models break the stationary-target assumption}
\label{sec:assumption}

When a world model does worse on previously seen data, there are two possible explanations, and its behavior alone does not reveal which one applies. Either it has lost knowledge that is still true, such as the position of the unmoved table, or it has correctly revised knowledge that the environment has since invalidated, such as the old position of the chair. Only the first is a learning failure, yet current evaluation reports both in the same way.

The continual world-model literature has largely not been required to confront this, because its benchmarks are constructed such that the second case cannot arise. Continual-Dreamer \citep{kessler2023effectiveness} is evaluated on Minigrid and Minihack task sequences, and WMAR \citep{yang2024augmenting} on Procgen and Atari. In both benchmarks the ground truth within a task never changes, so any later drop in performance is genuine forgetting. Some work states this assumption explicitly: DRAGO \citep{fu2025knowledge} studies task sequences that share the same dynamics and differ in their rewards. This is a reasonable simplification, but it is exactly the setting in which knowledge can never go out of date.

When environments do change, current world models adapt poorly. \citet{wan2022towards} show that PlaNet and DreamerV2 struggle to adapt to local changes, and trace this to a dilemma: a large replay buffer keeps outdated data that interferes with learning the change, while a small buffer forgets what is still true. Replay-level remedies discard stale data \citep{rahimikalahroudi2023replay, yang2026changepoint}, but they treat everything the model knows in the same way. What is missing is a formulation that separates what should be revised from what must survive.

\section{Stratification by invariance timescale}
\label{sec:strata}

We propose to organize world-model knowledge by how long it stays true, which we call its \emph{invariance timescale} $\tau$: the expected time until the environment makes a piece of knowledge false. $\tau$ is a property of the environment, not of the model; this is what distinguishes it from multi-timescale memory, where the timescales belong to the model \citep{behrouz2025nested, behrouz2025titans, jang2026musix}. In the household example, the laws of physics have $\tau=\infty$ (their form; parameters such as friction or mass can change) while the position of the chair has a short $\tau$; Table~\ref{tab:strata} groups knowledge into four such strata. An invariant must hold in every environment the model may be deployed in, not only in those seen so far, whereas a slowly changing regularity differs between environments. In a benchmark, $\tau$ is known by construction, because the designer decides how often each kind of knowledge changes; in deployment, it must be estimated from how often the model's predictions turn out to be wrong, for example by change-point detection~\citep{adams2007bayesian}.

\begin{table}[t]
\caption{Strata of world-model knowledge by invariance timescale $\tau$, with examples, required behavior, and how differential retention measures each (Sec.~\ref{sec:protocol}).}
\label{tab:strata}
\centering
\footnotesize
\setlength{\tabcolsep}{4pt}
\newcolumntype{L}[1]{>{\raggedright\arraybackslash}p{#1}}
\begin{tabularx}{\textwidth}{@{}L{2.5cm}>{\raggedright\arraybackslash}XL{2.7cm}L{2.9cm}@{}}
\toprule
\textbf{Stratum} & \textbf{Examples} & \textbf{Required behavior} & \textbf{Measured by} \\
\midrule
Invariants ($\tau=\infty$) & form of physical laws, object permanence, geometric consistency, action--effect structure & refined, never degraded & probes every round \\[2pt]
Environment regularities (long) & room layout, road network, friction, mass, the agent's body & revised on accumulated evidence & latency; not revised by one-off events \\[2pt]
Instance configuration (short) & object positions, door states, lighting & revised promptly & latency; never-revised count \\[2pt]
Transient state (immediate) & who is moving where, momentary occlusion & kept in current state, not in weights & not measured \\
\bottomrule
\end{tabularx}
\end{table}

The strata state how strongly each kind of knowledge should be protected, and they make evaluation well-posed: they determine which items enter the invariant probe set, which facts are timed for revision, and what counts as passing for each (Table~\ref{tab:strata}, last column). Note that $\tau=\infty$ describes the world, not the model. A model's belief about an invariant may be wrong and should be refined as evidence accumulates, but it must never get worse. A Kalman filter makes this trade-off: its update rate falls as $\tau$ grows and, for $\tau=\infty$, shrinks toward zero without reaching it \citep{kalman1960new}.

Existing consolidation methods do not follow this ordering, because they decide what to protect from the model's past rather than from how long knowledge stays true: elastic weight consolidation (EWC) \citep{kirkpatrick2017overcoming} protects parameters in proportion to their importance for earlier tasks, replay \citep{rolnick2019experience} keeps training on stored past experience, and distillation \citep{li2018learning} keeps the new model's outputs close to the old model's. Importance, however, is not the same as persistence: a parameter can be important for an earlier task and still encode a fact that the environment has since invalidated, and EWC then protects it most strongly. Current practice therefore fails in both directions. Standard pipelines do not check whether invariants survived an update, so the top stratum is under-protected, while a replay buffer that stores the old layout of a room keeps reasserting it after the room has changed, so the lower strata are over-protected. \citet{ashrafee2026holistic} address the second problem in classification by replacing outdated exemplars, and \citet{rahimikalahroudi2023replay} address it for world models by evicting stale transitions; the first problem, protecting invariants, has no counterpart in either setting.

\section{What current evaluation cannot measure}
\label{sec:metrics}

Standard continual-learning metrics such as backward transfer \citep{lopezpaz2017gradient} and the forgetting measure \citep{chaudhry2018riemannian} evaluate a model on an earlier task before and after later learning, and count any decrease as forgetting. We show that, in a changing environment, this cannot separate the two cases of Sec.~\ref{sec:assumption}. Consider the household robot, evaluated on its original test data after the chair has been moved. A model that correctly revises the chair and keeps the table loses one item, the chair; a model that keeps the stale chair but forgets the table also loses one item, the table. The metric gives both the same score, although the first behaves correctly and the second has failed twice, and a model that is never updated loses nothing and receives the best score.

This perverse optimum is harmless in a stationary benchmark, because a frozen model also fails on later tasks. In a changing environment it is not: apart from the rare case where new learning improves old tasks, every correct revision counts as forgetting, so the metric favors models that revise as little as possible, and the frozen model, the very failure that continual adaptation is meant to remedy, scores best.

Refreshing the labels, which repairs the metric in classification and textual factuality, is not enough for world models: a model can fit the updated labels of a rearranged room while degrading its representation of object permanence, because test data drawn from the task distribution rarely exercises object permanence directly.

\section{Differential retention}
\label{sec:protocol}

We propose \emph{differential retention}, which reports two quantities separately: \emph{invariant retention}, whether invariants survive adaptation, and \emph{revision latency}, how quickly changed facts are revised.

\textbf{(a) Invariant retention, measured by regression testing.} We fix a set of probes, small tests that each check one invariant, drawn from the $\tau=\infty$ stratum and held out from the adaptation data; for example, a probe might hide a ball behind the couch and check that the model still predicts the ball is there. We re-run the probes after every adaptation round, because a single evaluation at the end cannot show which update damaged an invariant. Existing physical probes test frozen video generators \citep{upadhyay2026worldbench, lin2026phyground}; for action-conditioned or latent world models, probes can instead read out the model's surprise at impossible events \citep{riochet2022intphys, garrido2025intuitive}, and similar probes can test geometric and causal structure. Alongside these probes, we re-test lower-stratum facts that the environment did not change, such as the position of the unmoved table. To our knowledge, such probes have not been applied as a regression suite across an adaptation stream.

\textbf{(b) Revision latency.} For each fact that the environment changes, we count how many observations of it the model needs before its predictions favor the new outcome over the old one on held-out queries, such as how many times the robot must see the chair at the table, and we count separately the facts it never revises. This generalizes the LoCA regret \citep{vanseijen2020loca} from a single local change to every changed fact.

Neither quantity is informative on its own, as Figure~\ref{fig:main}(b) illustrates: a frozen model loses no invariants but never revises, and a model that adapts by overwriting revises quickly but damages its invariants. We therefore report the two quantities separately, because a single score would hide which failure occurred, as it does for adaptive world models judged only by planning success \citep{wang2026adajepa}.

\section{Limitations and open questions}
\label{sec:limits}

This work contributes a framing and an evaluation protocol rather than experimental results, and it leaves four questions open. First, the strata in Table~\ref{tab:strata} are a starting proposal with domain-dependent boundaries; in benchmarks $\tau$ is set by design, but estimating it reliably in deployment, especially for slowly changing regularities, remains open. Second, mature probes exist mainly for physical invariants in generative video models; probes for geometric and causal structure still need to be developed and validated. Third, recurring knowledge, such as a door that is open on some days and closed on others, should be suspended and later reinstated rather than overwritten \citep{yang2026characterizing}. Finally, if invariants are entangled with other knowledge inside a learned world model, protecting them separately may require architectural support.

\begin{ack}
This research is partially supported by Adobe, Amazon, NVIDIA and Sesame.
\end{ack}

\bibliographystyle{plainnat}
\bibliography{refs}

\clearpage
\appendix

\section{Related work}
\label{app:related}

This appendix positions differential retention against the closest lines of work.

\textbf{Continual world models.} Continual-Dreamer \citep{kessler2023effectiveness}, WMAR \citep{yang2024augmenting} and DRAGO \citep{fu2025knowledge} are evaluated on task sequences whose dynamics stay fixed within each task, although \citet{kessler2023effectiveness} also show, in an appendix, that past data can interfere with learning when goals change. Other work studies what happens when the environment does change. The LoCA regret \citep{vanseijen2020loca} measures how quickly a model-based agent adapts to a local change in reward, at the level of the policy, and \citet{wan2022towards} show that PlaNet and DreamerV2 face a dilemma between interference from stale replay and forgetting. Retention itself is not the bottleneck: DreamerV3 with an unbounded replay buffer keeps essentially all measurable knowledge of earlier tasks \citep{nijjer2026world}, which in a changing environment would include dynamics that are no longer true. Several methods address this by discarding stale data: local forgetting evicts stale transitions near new ones \citep{rahimikalahroudi2023replay}, changepoint-aware world models flush stale replay after a detected dynamics shift \citep{yang2026changepoint}, and \citet{yang2026characterizing} show that discarding stale data helps after permanent shifts but hurts when earlier dynamics recur. These works act on the replay buffer and treat all knowledge alike. The shifts they study, such as changed gravity or actuator gain, are changes of physical parameters, which Table~\ref{tab:strata} places among regularities. None of them separates a stratum that must never be degraded, or tests whether that stratum survived adaptation.

\textbf{Changing ground truth and knowledge updating.} Concept drift methods handle shifting data distributions \citep{cabreramartin2026evolving}, for example by replacing outdated exemplars \citep{ashrafee2026holistic}. In language models, \citet{jiang2026benchmarks} measure how often an up-to-date answer is marked wrong by an outdated label, \citet{dhingra2022time} group facts by how often they change, and continual knowledge learning \citep{jang2022towards} separates time-invariant, outdated and new knowledge, combining forgetting and updating into a single ratio. In these settings, invariant knowledge consists of discrete facts that can be queried one at a time. In a world model, it is structure that underlies every prediction; differential retention probes this structure directly and reports retention and revision separately rather than as one ratio. Knowledge editing evaluates whether an edit succeeds and whether unrelated knowledge is preserved \citep{meng2022locating, yao2023editing}; these are the single-edit counterparts of revision latency and unchanged-fact retention.

\textbf{Evaluation metrics and regression testing.} \citet{diazrodriguez2018dont} argued that forgetting alone is insufficient for evaluating continual learners and proposed further criteria, and \citet{kumar2025continual} note that performance on previous tasks may overemphasize remembering; neither asks whether a drop in performance was warranted. In classification, the negative flip rate counts samples that an updated model gets wrong although its predecessor got them right. It ``can be made zero trivially by copying the old model'' \citep{yan2021positive}, which is the frozen-model optimum of Sec.~\ref{sec:metrics} in another form. Invariant regression testing applies the same idea to the $\tau=\infty$ stratum, where the correct answers never change and the criterion is therefore valid.

\textbf{Timescales of memory.} Fast and slow components appear in multi-timescale memory \citep{behrouz2025nested, behrouz2025titans}, in permanent and transient value functions for continual reinforcement learning \citep{anand2023prediction}, and in MuSix \citep{jang2026musix}, which gives each scale of a mixture of world models its own forgetting rate. MuSix places the models suited to low-level physical dynamics in its lowest-scale group, the one that forgets fastest, whereas in our stratification the form of physical laws is the most protected. In all of these works, the timescales belong to the model's memory. In ours, they are set by how fast the ground truth moves, they include a tier that is never degraded, and they determine what is measured. Only the rate at which ground truth moves can tell whether a change was correct: a fast-updating module that revises a physical invariant is working as designed, yet it is still wrong, because the invariant did not change.

\textbf{Evaluating and adapting world models.} WorldBench \citep{upadhyay2026worldbench} and PhyGround \citep{lin2026phyground} test individual physical concepts and laws in frozen video models. AdaJEPA \citep{wang2026adajepa} updates a latent world model online during planning and recovers under visual and dynamics shifts, as measured by planning success; planning success alone, however, cannot tell whether such recovery came from correctly revising an appearance prior or from degrading the dynamics into a form that happens to score well. Loss of plasticity \citep{dohare2024loss} constrains any continual scheme.

\textbf{Formal guarantees.} LeJEPA proves that its representation cannot collapse \citep{balestriero2025lejepa}, a guarantee that LeVJEPA carries over to video \citep{kuhn2026levjepa}; this says what a representation must \textit{not} do, not what it must keep. A structure is guaranteed to survive updates only when it is built into the architecture, as in networks with built-in symmetries or energy conservation \citep{cohen2016group, greydanus2019hamiltonian}, which is one route to the architectural support discussed in Sec.~\ref{sec:limits}.

\section{Assigning knowledge to strata}
\label{app:strata}

This appendix describes, without formal notation, what counts as knowledge, how it is assigned to a stratum, and why protection should grow with $\tau$.

\textbf{What counts as knowledge.} By knowledge we mean what a world model carries from one episode to the next, in its weights or in a persistent memory: the form of the dynamics, such as how unsupported objects fall, and facts about the environment, such as the layout of a room or where the chair is. Transient state, such as a person walking through the room, is tracked within an episode by the model's current state and is not stored as knowledge (Table~\ref{tab:strata}). A change in the environment makes a piece of knowledge false when the predictions that depend on it, such as what the robot expects to see when it looks at the corner by the window, are no longer correct. The invariance timescale $\tau$ is the expected time until this happens: if the chair is moved about once a week, $\tau$ for its position is about a week.

\textbf{Invariants and slow regularities.} An invariant must hold in every environment the model may be deployed in, for example every home a household robot will work in, and not merely have gone unchanged so far; this is invariance across environments in the sense of invariant risk minimization \citep{arjovsky2019invariant}. Which invariants hold therefore depends on where the model is deployed. A regularity, such as the layout of a room, stays fixed within one environment for a long time but differs between environments. Within physics, only the form of a law is invariant: its parameters, such as friction or mass, can change, as can the agent's own body, and they belong to a lower stratum, as in hidden-parameter models of dynamics \citep{doshivelez2016hidden}. Conversely, the top stratum is not only physics: object permanence, geometric consistency and causal action--effect structure belong there too, and physics is simply the part for which probes are most developed. The four strata divide a continuous $\tau$ into the classes that call for different behavior and different measurement.

\textbf{$\tau$ in benchmarks and in deployment.} In a benchmark, $\tau$ is known by construction. The designer decides which properties are the same in all generated environments, which are regenerated every so many episodes, and which change more often, so the stratum of every probe and every scored fact is known exactly and the evaluation in App.~\ref{app:protocol} needs no estimate of $\tau$. In deployment, $\tau$ must be estimated from how often the model's predictions turn out to be wrong. For example, change-point detection applied to the model's prediction errors \citep{adams2007bayesian} marks when the environment has changed, and the average interval between detected changes estimates $\tau$. From a short history, a true invariant cannot be told apart from a slowly changing regularity, so an item is treated as provisionally invariant until it is contradicted. Mistaking a slow regularity for an invariant slows its revision but does not prevent it, because invariants are consolidated rather than frozen, and repeated contradiction is itself evidence that the item belongs in a lower stratum.

\textbf{Recurring knowledge.} Some facts change back and forth, such as a door that is open on some days and closed on others. Such a fact is better represented as context-dependent: the model keeps both values and infers which one currently applies, so the old value is suspended rather than overwritten~\citep{yang2026characterizing}. Revision latency then measures how quickly the model switches between them.

\textbf{Why protection should grow with $\tau$.} Optimal estimation gives the same ordering in the simplest case. Consider a single quantity, such as the position of the chair along one axis, that drifts over time and is observed with noise, and that is tracked by a Kalman filter \citep{kalman1960new}, the optimal estimator when drift and noise are Gaussian. Each observation moves the estimate by a fraction of the surprise, the difference between what was observed and what was expected. This fraction is the filter's learning rate, and the remainder is the protection it gives to what it already knows. The more slowly the quantity drifts relative to the observation noise, the smaller the learning rate, so protection grows with how long the quantity stays the same. For a quantity that never changes, the learning rate keeps shrinking as observations accumulate but never reaches zero: the estimate becomes an average of everything observed so far, which keeps improving with evidence and can still correct an early mistake, but responds less and less to any single observation. Two caveats apply. A single noisy observation can still move a correct estimate slightly away from the truth, so the filter motivates ``refined, never degraded'' as a target rather than guaranteeing it. And the filter's timescale depends on how noisy the observations are as well as on how fast the world changes, so it is related to $\tau$ but not identical to it. The same ordering, smaller steps for what changes more slowly, holds for gradient-based learners whose targets drift \citep{kumar2025continual}. It also implies that even an optimal learner takes longer to revise slowly changing knowledge, so targets for revision latency should be set per stratum rather than once for all knowledge.

\section{Implementing differential retention}
\label{app:protocol}

This appendix describes how differential retention (Sec.~\ref{sec:protocol}) can be implemented. The adaptation stream is divided into rounds, and every measurement is repeated after each round.

\textbf{Invariant retention.} Each probe in the held-out invariant set has a pass criterion: depending on the probe, a prediction error below a tolerance or, for violation-of-expectation probes, higher surprise on the impossible outcome than on the possible one (App.~\ref{app:probes}). After every round we report the fraction of probes passed, separately for each concept such as object permanence; its lowest value over the stream; and the \emph{negative flips}, probes that the model passed before adaptation but fails afterwards~\citep{yan2021positive}. Forgetting can be temporary and recover before a round ends \citep{delange2023continual}, so the probes should also be run within rounds. Each probe checks correctness against the world rather than agreement with the model before adaptation, so a model that corrects a mislearned invariant improves its score instead of being penalized for changing.

\textbf{Retention of unchanged facts.} The same procedure is applied to held-out facts from the lower strata that the environment did not change, such as the position of the unmoved table. This catches a failure that the other measurements miss: a model that relocates the moved chair but forgets where the unmoved table stands passes the invariant probes and revises the chair quickly, and fails only here. In knowledge editing, the same requirement is known as locality \citep{meng2022locating, yao2023editing}.

\textbf{Revision latency.} For each scripted change, such as the chair moving from the window to the table, we ask on held-out queries whether the model now prefers the new outcome over the old one by a clear margin, that is, whether it assigns the new outcome a clearly lower prediction error. The latency is the number of observations of the changed region after which this preference holds for several consecutive observations, so that a single lucky prediction does not count. Latency counts observations of the changed region, not training steps. The queries are held out rather than taken from the next observation, because a model that simply copies its last input would otherwise appear to revise instantly, and they are posed in a new episode, so that tracking the chair within the current episode is not mistaken for revising what the model knows. A change that is still not revised when the next change arrives or the stream ends is counted as never revised, while a change that arrives too close to the end to be judged is reported separately rather than counted as a failure. We report the median latency per stratum and the count of changes that are never revised.

\textbf{False revisions.} A regularity should not be revised because of a single one-off event, such as a corridor that is blocked for a moment. We therefore also apply scripted perturbations that later revert, and report the fraction of regularity items whose predictions flip in response. Detecting changes faster always comes at the cost of more false alarms, so latencies are comparable between learners only at a stated false-revision rate.

\textbf{Reporting.} We report invariant retention for the top stratum, and retention of unchanged facts and revision latency for each lower stratum, together with the never-revised count and the false-revision rate, without combining them into a single score. Two reference learners anchor the results: the frozen model, which marks the top-right corner of Figure~\ref{fig:main}(b), and unconstrained fine-tuning.

\textbf{A minimal benchmark.} A benchmark needs a family of environments in which each stratum changes on a known schedule: the simulator's physics is the same in all environments (invariants); the layout is regenerated every so many episodes (regularities); door states and object positions change more often (instance configuration); and other agents move within episodes (transient state). Platforms with controllable factors of variation, such as stable-worldmodel \citep{maes2026stable}, already provide much of this. The model is trained continually on this stream, and after every round one runs the invariant probes, the unchanged facts, and the revision queries for each scripted change.

\section{Probes beyond physics}
\label{app:probes}

Differential retention needs a probe for each kind of invariant, not only for physics. This appendix sketches such probes. Existing video benchmarks such as WorldBench \citep{upadhyay2026worldbench} and PhyGround \citep{lin2026phyground} test generative video models: they condition the model on a short video or a first frame and score the video it generates, without any actions. Action-conditioned or latent world models, such as Dreamer-style or JEPA-style models, need probes of a different form. These probes can read out the model's own prediction error, for example whether it is more surprised by an impossible event than by a possible one, or they can be built inside the training simulator, where the correct outcome is known by construction. Table~\ref{tab:probes} gives one probe design for each kind of invariant, with household examples. Whatever their form, probes should only read the model and should be simple: training a probe or an adapter on top of the model can distort the very structure being measured \citep{interno2026observer}.

\begin{table}[h]
\caption{Probe designs for each kind of invariant, with household examples. All can be run on action-conditioned or latent world models by reading out the model's prediction error; the last column lists existing resources.}
\label{tab:probes}
\centering
\footnotesize
\setlength{\tabcolsep}{4pt}
\begin{tabularx}{\textwidth}{@{}>{\raggedright\arraybackslash}p{2.3cm}>{\raggedright\arraybackslash}X>{\raggedright\arraybackslash}p{3.5cm}>{\raggedright\arraybackslash}p{3.0cm}@{}}
\toprule
\textbf{Invariant} & \textbf{Probe} & \textbf{Pass criterion} & \textbf{Existing resources} \\
\midrule
Physical laws (form) & show a possible and an impossible continuation of the same scene, e.g.\ a dropped cup that falls or one that floats (violation of expectation) & higher surprise on the impossible continuation & \citet{riochet2022intphys}; \citet{garrido2025intuitive}; WorldBench, PhyGround (video) \\[2pt]
Object permanence & hide an object behind an occluder, then remove the occluder, e.g.\ a ball behind the couch & object predicted to still be there; surprise if it has vanished & \citet{riochet2022intphys}; \citet{garrido2025intuitive} \\[2pt]
Impenetrability & move into a wall or a solid object & predicted state does not pass through it & built in the simulator \\[2pt]
Geometric consistency & loop closure: a sequence of moves that returns to the start, e.g.\ walking around the table; several views of one scene & predicted end state matches the start; views agree with each other & MEt3R \citep{asim2025met3r} (generated video) \\[2pt]
Causal action--effect & from one state, compare two different actions, e.g.\ pushing or pulling the door; also the null action & predicted difference matches the simulator; the null action leaves a static scene unchanged & CLEVRER \citep{yi2020clevrer} (video question answering) \\
\bottomrule
\end{tabularx}
\end{table}

\end{document}